\documentclass{bmvc2k}
\usepackage{amssymb}
\usepackage{amsmath,amsfonts}
\usepackage{amsopn}
\usepackage{bm} 
\usepackage{multirow}

\newlength\savewidth\newcommand\shline{\noalign{\global\savewidth\arrayrulewidth\global\arrayrulewidth1pt}\hline\noalign{\global\arrayrulewidth\savewidth}}

\newcommand{\vct}[1]{\boldsymbol{#1}} 

\newcommand{\ProbOpr}[1]{\mathbb{#1}}

\newcommand{\expect}[2]{%
\ifthenelse{\equal{#2}{}}{\ProbOpr{E}_{#1}}
{\ifthenelse{\equal{#1}{}}{\ProbOpr{E}\left[#2\right]}{\ProbOpr{E}_{#1}\left[#2\right]}}} 
\newcommand{\var}[2]{%
\ifthenelse{\equal{#2}{}}{\ProbOpr{VAR}_{#1}}
{\ifthenelse{\equal{#1}{}}{\ProbOpr{VAR}\left[#2\right]}{\ProbOpr{VAR}_{#1}\left[#2\right]}}} 

\newcommand{\vtheta}{\vct{\theta}}

\newcommand{\vx}{{\vct{x}}}
\newcommand{\vy}{\vct{y}}

\newcommand{\vv}{\vct{v}}
\newcommand{\vw}{\vct{w}}

\newcommand{\eat}[1]{}

\usepackage{comment}
\usepackage{floatrow}
\usepackage{multirow}
\usepackage{rotating} 
\usepackage{bbm}
\usepackage{dsfont}
\usepackage{xcolor}
\usepackage{wrapfig}

\newfloatcommand{capbtabbox}{table}[][\FBwidth]

\title{Thoracic Disease Identification and Localization using Distance Learning and Region Verification}

\addauthor{Cheng Zhang}{zhang.7804@osu.edu}{1}
\addauthor{Francine Chen}{francine@acm.org}{2}
\addauthor{Yan-Ying Chen}{yanying@gmail.com}{2}

\addinstitution{
 The Ohio State University \\
 Columbus, OH, USA
}
\addinstitution{
 FX Palo Alto Laboratory \\
 Palo Alto, CA, USA
}

\runninghead{Zhang, Chen, Chen}{Thoracic Disease Identification and Localization}

\renewcommand{\paragraph}[1]{\vspace{1ex}\noindent\textbf{#1}}
\def\ie{\emph{i.e}\bmvaOneDot}

\def\etal{\emph{et al}\bmvaOneDot}
\newcommand{\cmmnt}[1]{}
\begin{document}
\maketitle
\begin{abstract}
The identification and localization of diseases in medical images using deep learning models have recently attracted significant interest. 
Existing methods only consider training the networks with each image independently and most leverage an activation map for disease localization.
In this paper, we propose an alternative approach that learns discriminative features among triplets of images and cyclically trains on region features to verify whether attentive regions contain information indicative of a disease.
Concretely, we adapt a distance learning framework for multi-label disease classification to differentiate subtle disease features.
Additionally, we feed back the features of the predicted class-specific regions to a separate classifier during training to better verify the localized diseases.
Our model can achieve state-of-the-art classification performance on the challenging ChestX-ray14 dataset, and our ablation studies indicate that both distance learning and region verification contribute to overall classification performance.
Moreover, the distance learning and region verification modules can capture essential information for better localization than baseline models without these modules. 
\end{abstract}
\section{Introduction}
\label{sec:intro}
\begin{figure}[t]
    \centerline{\includegraphics[width=1\linewidth]{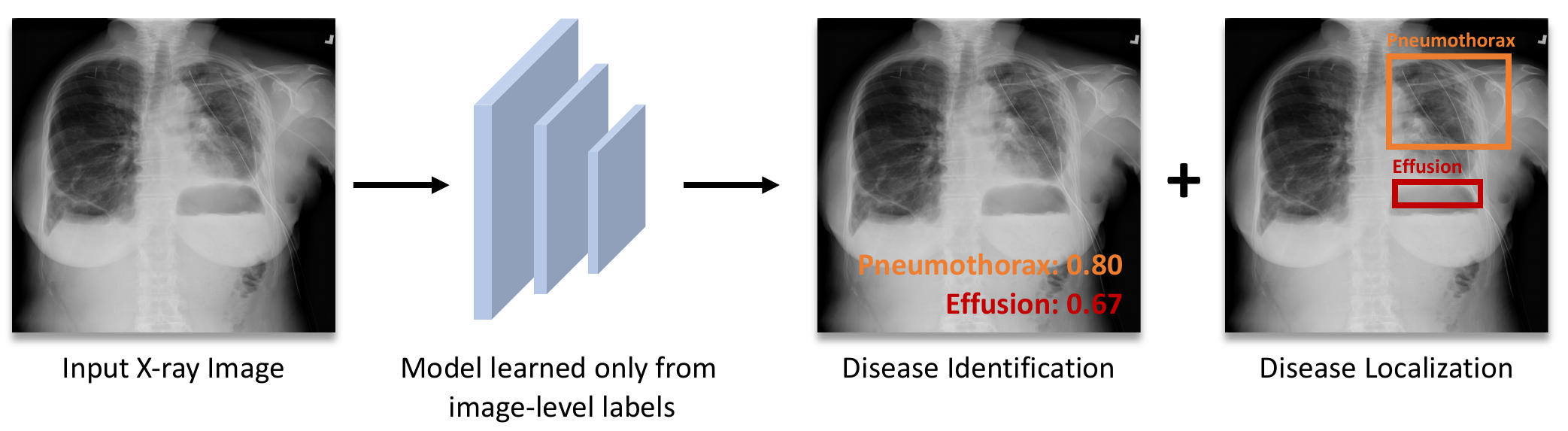}}
    \caption{\small \textbf{Overview of thoracic disease identification and localization} trained on chest X-rays and image-level disease labels. Given a test image, the model predicts how likely the diseases occur in the image and localizes their corresponding bounding boxes. Multiple diseases could co-exist in an image.}
    \label{fig:1}
\end{figure}

Radiography has been widely adopted for detecting a number of thoracic diseases. However, detecting diseases in X-ray images requires the expert knowledge of radiologists, who are overburdened and often must quickly review each image. Further, the location of an identified disease is generally not annotated and so may be unclear to another doctor reviewing the X-ray. Several datasets have been released for disease classification from chest X-ray images, including \cite{wang2017chestx,johnson2019mimic,irvin2019chexpert,bustos2019padchest}, and
the use of deep learning models in combination with the datasets has resulted in much progress~\cite{chen2019develop,wang2017chestx,wang2018tienet,rajpurkar2018deep,li2018thoracic,cai2018iterative}. Nevertheless, the identification and localization of thoracic diseases are still challenging due to subtle inter-disease differences and large intra-disease variations across different subjects and regions. 

The only chest X-ray dataset with disease bounding boxes, ChestX-ray14~\cite{wang2017chestx}, has boxes only for the test partition. Under this constraint, it is difficult to apply supervised learning to disease localization, and two popular weakly supervised approaches have been proposed for disease identification and localization: CAM-based (class activation map)~\cite{zhou2016learning} and MIL-based (multi-instance learning). Both embed an X-ray image using a pre-trained image embedding network such as ResNet~\cite{he2016deep} or DenseNet~\cite{huang2017densely}. Under CAM-based, the computed features are used to train a multi-label classifier on disease labels; and localization is performed based on the class activations in the embedding network~\cite{wang2017chestx,rajpurkar2017chexnet}. In MIL-based, a grid is formed from the embedded features and each element of the grid is classified as to which, if any, diseases occur, indicating which diseases occur in the image; and localization is performed by combining the grid classifications~\cite{liu2019align,li2018thoracic}. 

In this paper, we focus on multi-label, weakly supervised thoracic disease identification (we will use the terms \emph{identification} and \emph{classification} interchangeably) and localization in chest X-rays using a weakly supervised learning approach (see \autoref{fig:1}) . 
Different from existing methods that learn a neural network model on each image independently, we leverage distance learning~\cite{schroff2015facenet,balntas2016learning,hermans2017defense,zheng2019hardness,zhang2016siamese,jiang2018learning} to learn sufficient feature representations to tackle multi-label disease classification. In particular, we exploit triplets of images as the inputs to drive the similarity metric to be small for the pairs of images with similar diseases, and large for the pairs with different ones. Furthermore, we propose a region verification module that feeds back the class-specific attentive regions to verify those regions important to disease classification. The size of the regions vary, in contrast to the full images used with previous CAM-based approaches and the smaller contexts from the  grids used in the MIL-based approaches. Finally, with the well-learned disease features, our model can be used for disease localization by incorporating weakly-supervised object detection methods such as CAM.

Evaluation on the ChestX-ray14 dataset shows that our model achieves state-of-the-art classification performance and consistently outperforms CAM and MIL baselines for multiple diseases.  Ablation studies indicate that both distance learning and region verification help classification performance, with the strongest contribution from region verification. Moreover, our experiments also indicate that both modules improve disease localization over baseline models without these modules, strengthening the support to the improvement on classification since the diseases are better localized by verified regional features. In conclusion, our contributions include: (1) design an end-to-end framework for training multi-label disease classification and weakly supervised localization simultaneously; (2) extend distance learning to multi-disease medical images; (3) propose region verification to align disease classification of a whole image and the local context surrounding the disease. 
\section{Related Work}
\label{sec:related}
\paragraph{Disease identification.}
Recently, significant progress on X-ray classification and detection has been made using deep neural networks. Bar~\etal.~\cite{bar2015chest} presented an early examination of the strength of deep learning approaches for pathology detection in chest radiographs.  A number of works developed thoracic disease classification models on the more comprehensive ChestX-ray14 dataset~\cite{wang2017chestx,rajpurkar2018deep,rajpurkar2017chexnet,li2018thoracic,guan2020thorax,cai2018iterative,liu2019align}. Wang \etal.~\cite{wang2017chestx} and Rajpurkar \etal.~\cite{rajpurkar2018deep,rajpurkar2017chexnet} explored the use of deep learning models for disease classification. Li~\etal.~\cite{li2018thoracic} proposed to unify the training of image-level and box-level labels in one framework with a customized MIL loss where disease classification is performed on a grid over the image.  Several works applied an attention mechanism to chest X-ray analysis~\cite{ma2019multi,guan2020thorax,cai2018iterative}. For example, Guan~\etal~\cite{guan2020thorax} designed an attention guided two-branch network for thorax disease classification, which helps amplify the \emph{high activation} regions. Cai~\etal.~\cite{cai2018iterative} presented an attention mining strategy to improve the model's sensitivity or saliency to disease patterns. Liu~\etal.~\cite{liu2019align} proposed an alternative method for computing attention based on the difference between the input image 
and an image without any disease (a ``negative image''), but did not employ the pair in distance learning. Different from the above methods that only consider single/pair-level image information and class-agnostic attention features, we take advantages of triplets of images with \emph{distance learning} for \emph{multi-label} disease classification and cyclically train~\cite{ma2019learning} the networks with \emph{class-specific} attentive region features.

\paragraph{Weakly-supervised disease localization.}
Fully supervised detection methods have achieved great success in identifying objects when trained on a vast number of bounding box annotations ~\cite{ren2015faster,girshick2014rich,liu2016ssd,redmon2016you}. However, such annotations are very expensive to create manually for  medical images, requiring busy radiologists to do the annotation. Therefore, weakly supervised approaches to object localization using disease labels for an image are commonly used.  One popular approach when classification is performed over the full image is to estimate disease locations based on class activation maps (CAM)~\cite{zhou2016learning} or gradient-weighted class activation maps (Grad-CAM)~\cite{selvaraju2017grad}.
Zhou~\etal.~\cite{zhou2016learning} utilized a global average pooling layer for neural networks to generate CAM that are used to localize objects. 
The use of CAM-based methods was employed by \cite{wang2017chestx,rajpurkar2018deep,rajpurkar2017chexnet,li2018thoracic,guan2020thorax,cai2018iterative,he2018fast}. Often, the thresholded CAM activations are referred to as ``attention'' or ``attentive regions''.
In another approach employing MIL as used by Li \etal.~\cite{li2018thoracic} and Liu~\etal.~\cite{liu2019align}, the cells of the image grid for which diseases are predicted are combined to predict disease locations. 
Most prior works leverage attentive activation maps to localize disease regions while neglecting the domain gap between classification and localization results. We propose region verification which feeds back CAM-based region features to a local classifier to verify the disease localized by CAM. 
Our region verification module is class-specific and could be also incorporated with other weakly-supervised object localization methods. 

\paragraph{Distance learning.} 
Distance (metric) learning (DML) generally works with two types of data: \emph{pair-wise}  with must-link and cannot-link constraints, and \emph{triplet} constraints that contains a similar pair and a dissimilar pair. 
The triplet loss with semi-hard mining was introduced in~\cite{schroff2015facenet} to compute image embeddings for identifying faces and extended to improve performance or computation~\cite{balntas2016learning,hermans2017defense}.
While earlier distance learning tasks focused on re-identification~\cite{taha2020boosting,zheng2019hardness,liu2018learning} of one class of object per image, we investigate how distance learning can be extended to multi-label disease images. Instead of directly using the disease features extracted by the pre-trained deep model, we present a first attempt to employ triplet learning for a multi-label disease classification task.
\section{Approach}
\label{sec:approach}

\begin{figure*}[t]
    \centerline{\includegraphics[width=0.9\linewidth]{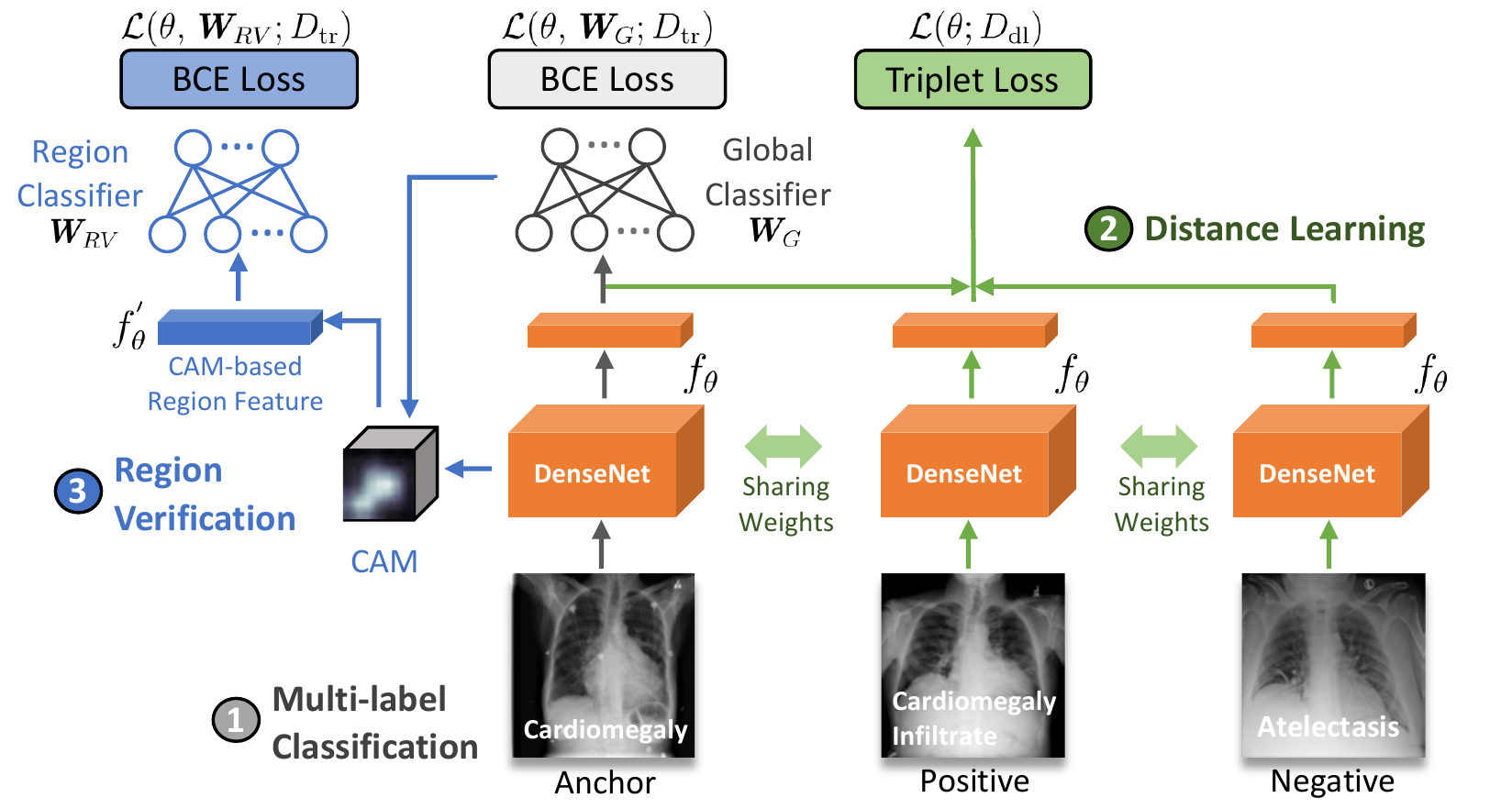}}
    \caption{\small \textbf{Framework of disease identification and localization.} Our model takes triplets of images as the inputs, \ie, anchor, positive, and negative for distance learning, and feeds back the CAM-based attentive feature to the region classifier for disease verification. The pipeline is trained end-to-end.} 
    \label{fig:framework}
  \vspace*{-3mm}
\end{figure*}

An overview of our proposed training framework is shown in \autoref{fig:framework}. Our model takes triplets of images as the input and consists of three parts: (1) multi-label classification of the anchor image with a conventionally trained binary cross entropy (BCE) loss (2) distance learning with triplet loss and multi-label hard example mining and (3) region verification of attentive local areas with BCE loss. We jointly train the three loss functions end-to-end. We next describe each component.

\subsection{Multi-label classification}
A disease identification model takes an input X-ray image $\vx$ and predicts a confidence score for each disease $c$ by
\begin{align}
p(y_c|\vx) = \sigma (\vw_c^\top f_{\vtheta} (\vx)),
\end{align}
where $f_{\vtheta}(\cdot)$ is an embedding network parameterized by $\vtheta$, and $\vw_c$ is the linear classifier of disease type $c$. Here we formulate disease identification as a multi-label classification problem. Given a training set $D_\text{tr} = \{\vx_n, \vy_n\}_{n=1}^N$, in which $\vx_n$ is the input image to be classified and $\vy_n\in\Delta^C$ is a vector on the $C$-dimensional simplex. $C$ is the size of the disease types. We train the embedding network using empirical risk minimization (ERM) with BCE loss:
\begin{align}
    \mathcal{L} (\vtheta,\emph{\textbf{W}}_G; D_\text{tr}) = - \sum_n \sum_c \vy_{n,c}\log p(y_{n,c}|\vx_n) + (1 - \vy_{n,c}) \log (1 - p(y_{n,c}|\vx_n)),
    \label{eq:global}
\end{align}
where the global branch classifier $\emph{\textbf{W}}_{G}=\{\vw_c\}_{c=1}^{C}$ and stochastic gradient descent (SGD) optimization is applied with uniformly sampled instances from $D_\text{tr}$.

\subsection{Multi-label distance learning with triplet loss}
\label{sec:triplet}
We adapt a multi-label classification system to use a triplet loss to encourage images with at least one common disease to be close together in feature space, and images without any common disease to be far apart in feature space. A triplet loss minimizes the distance between the image to be classified, or \emph{anchor}, and another image of the same type.
We examined two cases as \emph{positive} examples: (1) only images with an identical set of labels and (2) images from case (1) and also images with partial label matches.
\emph{Negative} examples were defined as images for which the intersection of disease labels with the anchor is null. Since small batches may not contain exact matches for some label combinations, we also proposed to pre-compute similarity on a randomly selected subset of exact match, partial match, or no match images, rather than by batch during training, leading to more efficient training.

Specifically, we consider a distance (triplet) learning constraint set $D_\text{dl}$ given as follows:
\begin{align}
    D_\text{dl} = \{(\vx_i, \vx^{+}_i, \vx^{-}_i) | (\vx_i, \vx^{+}_i) \in \mathcal{P}; (\vx_i, \vx^{-}_i) \in \mathcal{N}\},
\end{align}
where $i=1,2,\ldots,M$ ($M$ is the cardinality of the entire triplet set). $\mathcal{P}$ contains positive pairs and $\mathcal{N}$ includes negative pairs. We denote the similarity of two samples as $\ell_2$ distance between the feature embeddings and define the hinge loss for one triplet given the anchor $\vx_i$:
\begin{align}
    l(\vx_i, \vx^{+}_i, \vx^{-}_i) = [\left\| f_{\vtheta}(\vx_i), f_{\vtheta}(\vx_i^{+}) \right\| - \left\| f_{\vtheta}(\vx_i), f_{\vtheta}(\vx_i^{-}) \right\| + m]_{+},
\end{align}
where $m$ is a margin (0.5 in the experiments) that is enforced between positive and negative pairs. Therefore the triplet loss is  minimized over all possible triplets in the distance learning set $D_{\text{dl}}$, which can be computed as:
\begin{align}
    \mathcal{L} (\vtheta; D_\text{dl}) = \sum_{(\vx_i, \vx^{+}_i, \vx^{-}_i)\in D_\text{dl}} l(\vx_i, \vx^{+}_i, \vx^{-}_i).
    \label{eq:triplet}
\end{align}
\paragraph{Hard example mining.} The distance learning constraint set $D_\text{dl}$ is constructed by hard example mining,~\ie., selection of semi-hard training examples relative to the anchor image, to make training a model more effective~\cite{schroff2015facenet}. 
To identify harder examples for training, we sort the set of positive images and the set of negative images for an anchor image by perceptual similarity to the anchor using a perceptual hash~\cite{zauner2010implementation,liu2019align}.
To reduce computation prior to sorting for identifying hard examples for an anchor, a randomly selected set of 1000 negative examples, and a randomly selected set of up to 500 positive examples were identified for each anchor. To give preference to exact matches when partial matches are used, we set 25\% of the  positive examples to be partial matches, while the remainder are exact matches. The examples for each anchor are randomly selected with a bias towards easier examples initially and the hardest examples after 10 epochs. 

\subsection{Region verification}
Intuitively, if the disease location is predicted incorrectly, the classification will likely be incorrect. To align disease localization and classification, we propose to feed the attentive region features into another region classifier for cyclical training, namely \emph{region verification} (RV). In the existing single-class CAM approach, weighted activations for a single class are used for localization. In contrast, in our multi-label setting, there can be multiple classes. To handle this, we compute class-dependent activations by projecting back the weights of the output layer separately for each class. We then merge the attentive regions of all identified classes (labeled for training, predicted for testing) in each image and zero out the activations of non-attentive regions to use as features in region verification. This also contrasts with related work that uses class-independent, high activation regions.

As shown in \autoref{fig:framework}, we use CAM~\cite{zhou2016learning} to identify the region, where the activations at the last layer of DenseNet are extracted with the weights of global image classifier:
\begin{align}
    M_c(a, b) = \sum_{k} w_{c}^{k} \cdot g_k(a,b),
    \label{eq:cam1}
\end{align}
where $w_{c}^{k}$ is the scalar weight of disease class $c$ for feature $k$. We use $g_k(a,b)$ to denote a value of 2-dimensional spatial point $(a, b)$ with feature $k$ in map $g_k$ from the final convolutional layer of DenseNet-121, where the sizes of $k$ and $g_k$ are 1024 and $1024\times7\times7$, respectively. 

To extract attentive region features, our class-specific activation map considers all feature maps associated to only ``positive'' diseases (labels) in the anchor image:
\begin{align}
    M(a, b) = \sum_{c} \mathbbm{1}[y_c=1] \cdot M_c(a, b).
    \label{eq:cam2}
\end{align}
Next, we normalize the activation map to [0, 1] over the difference between the maximum
and minumum value over all features. The locations of feature activations that
are greater than a fixed threshold (we empirically set to 0.8 in the experiments) are identified. Then a rectangular bounding box around the thresholded activations is computed and features outside the bounding box are zeroed. The resulting features are fed into the region classifier, which is a separate classifier randomly initialized. Finally, we optimize the region classifier and the shared embedding network by using a binary cross entropy loss as follows:
\begin{align}
    \mathcal{L} (\vtheta,\emph{\textbf{W}}_{RV}; D_\text{tr}) = - \sum_n \sum_c \vy_{n,c}\log \sigma (\vv_{c}^\top f_{\vtheta}' (\vx)) + (1 - \vy_{n,c}) \log (1 - \sigma (\vv_{c}^\top f_{\vtheta}' (\vx))),
    \label{eq:region}
\end{align}
where $\emph{\textbf{W}}_{RV}=\{\vv_{c}\}_{c=1}^{C}$ is the region verification classifier and $f_{\vtheta}'$ is the distilled CAM-based attentive region feature.

\subsection{Joint learning and inference}
We jointly optimize the parameters of the model. The gradients are computed from three losses. The main loss $\mathcal{L} (\vtheta,\emph{\textbf{W}}_G; D_\text{tr})$ refers to binary cross entropy loss associated with the whole image prediction from \autoref{eq:global}. We backpropagate this loss to optimize the parameters of the feature embedding network and global classifier $\emph{\textbf{W}}_G$. The region verification loss $\mathcal{L} (\vtheta,\emph{\textbf{W}}_{RV}; D_\text{tr})$ is another binary cross entropy loss for disease class predictions from attentive region features from \autoref{eq:region}. We use this loss to optimize the shared feature embedding parameters $\vtheta$ and region verification classifier $\emph{\textbf{W}}_{RV}$. By doing so, we further improve the model's ability to capture intra-image subtle disease features. Based on \autoref{eq:triplet}, $\mathcal{L} (\vtheta; D_\text{dl})$ is the triplet loss during learning of inter-image relationships from distance learning set $D_\text{dl}$. We obtain the final loss $\mathcal{L}_{total}$ by adding the three losses together:
\begin{align}
    \mathcal{L}_{total} (\vtheta,\emph{\textbf{W}}_G, \emph{\textbf{W}}_{RV}; D_\text{tr}) = \mathcal{L} (\vtheta,\emph{\textbf{W}}_G; D_\text{tr}) + \mathcal{L} (\vtheta,\emph{\textbf{W}}_{RV}; D_\text{tr}) + \mathcal{L} (\vtheta; D_\text{dl})
\end{align}

\paragraph{Disease identification.}
We average the decision values by considering both global and region classifiers to compute the confidence score for each disease:
\begin{align}
p_{total}(y_c|\vx) = \sigma (\vw_c^\top f_{\vtheta} (\vx) + \vv_c^\top f_{\vtheta}' (\vx)).
\end{align}

\paragraph{Disease localization.}
We average the weights of the global and region classifiers to compute the activation map for one disease $c$:
\begin{align}
    M_c(a, b) = \sum_{k} \frac{1}{2}(w_{c}^{k} + v_{c}^{k}) \cdot g_k(a,b).
    \label{eq:cam3}
\end{align}
We normalize the activation map $M_c(a, b)$ and apply a selected threshold to generate the predicted bounding box. Details and qualitative results can be found in the experiments.
\vspace*{-2mm}
\section{Experiments}
\label{sec:exp}

\subsection{Setup}
\paragraph{Dataset.}
We evaluate our approach on the NIH ChestX-ray14 dataset~\cite{wang2017chestx}, which consists of 112,120 frontal-view X-ray images of 30,805 patients with 14 disease labels and each image can have multi-labels, and we follow its official train/test split. In a subset of the test set used for evaluation of the localization task, 880 images are labeled with 984 bounding boxes for 8 diseases by board-certified radiologists.  We reserved 10\% of the images from the training set as validation based on patient ID, thus insuring that images of an individual patient are present in only one of the train, validation or test sets. Consequently, we split data into 77,821/8,703/25,596 images for training/validation/testing.

\paragraph{Evaluation metrics.}
We follow \cite{wang2017chestx,li2018thoracic,liu2019align} to evaluate our approach. 
For disease classification, we use AUC scores (the area under the ROC curve) to measure the performance. For disease localization, we evaluate the detected regions against the ground truth (GT) bounding boxes, using accuracy for a given intersection over union ratio (IoU). The localization results are only calculated for those 8 disease types with GT provided. A correct localization is defined when IoU > T, where T is the threshold.

\paragraph{Implementation details.}
We resize the original 3-channel X-rays from 1024 $\times$ 1024 to 224 $\times$ 224 for faster processing. We apply a 5 degree random rotation and horizontal flipping during training for data augmentation. The DenseNet-121~\cite{huang2017densely} model, pretrained with ImageNet, is used as the backbone because of its better performance on disease classification as shown in~\cite{rajpurkar2017chexnet}. For all models, we train for at most 30 epochs using SGD with Adam~\cite{kingma2014adam}. The initial learning rate is $10^{-3}$, which is divided by 10 after E epochs. We tune E on the validation set and choose the best model via validation performance. For disease localization, since no box annotations are available for validation, we apply the same threshold on CAM for each disease type using 10-fold cross-validation of the test data to predict bounding boxes and report the results on the localization set.

\vspace{-2mm}
\subsection{Disease identification}
\vspace{-2mm}
\paragraph{Main results.}
We compare our model's disease identification performance to four baselines as shown on the left side of \autoref{tab:result_cls} on the official test split of the ChestXray14 dataset.  We did not compare against Guan \etal~\cite{guan2020thorax} who tested on a different split. The left two \cite{wang2017chestx,rajpurkar2017chexnet} are whole image, CAM-based models and the right two \cite{liu2019align,li2018thoracic} are MIL-based models. Note that Li \etal~\cite{li2018thoracic} and CIA-Net~\cite{liu2019align} used a combination of 70\% unannotated data and 70\% annotated cross-validation test data for training. The other models, including ours, are trained only on unannotated data. The results show that our model, which uses region verification and distance learning, outperforms all baselines on average, and is the top-performing model for 10 of the 14 diseases, indicating the effectiveness of joint use of region verification and distance learning for improving disease identification performance. 

\begin{table}[]
\centering
    \begin{tabular}{c|cccc|c|cc}
Disease             & \cite{wang2017chestx}  & \cite{rajpurkar2017chexnet}$^{\ast}$ & \cite{li2018thoracic}$^{\diamond}$   & \cite{liu2019align}$^{\diamond}$ &  Ours & \color{gray}{w/o RV} & \color{gray}{w/o DL} \\
\shline
Atelectasis         & 0.700 & 0.755   & \underline{{0.80}} & 0.79 & \textbf{{0.845}} & \color{gray}{0.802} & \color{gray}{0.833}  \\
Cardiomegaly        & 0.810 & 0.867   & \underline{{0.87}} &  \underline{{0.87}}  & \textbf{{0.905}} & \color{gray}{0.856} & \color{gray}{0.887}\\
Effusion            & 0.759 & 0.815   & 0.87 & \textbf{{0.88}}& \underline{{0.877}}& \color{gray}{0.849} &  \color{gray}{0.861} \\
Infltration         & 0.661 & 0.694   & \underline{{0.70}} & 0.69 & \textbf{{0.817}}& \color{gray}{0.797} &   \color{gray}{0.799} \\
Mass                & 0.693 & 0.802   & \underline{{0.83}} & 0.81 & \textbf{{0.859}}& \color{gray}{0.826} &   \color{gray}{0.835} \\
Nodule              & 0.669 & 0.735   & \underline{{0.75}} & 0.73 & \textbf{{0.824}}& \color{gray}{0.752} &   \color{gray}{0.792} \\
Pneumonia           & 0.658 & 0.698   & 0.67 & \underline{{0.75}} & \textbf{{0.804}}& \color{gray}{0.751} &   \color{gray}{0.791} \\
Pneumothorax        & 0.799 & 0.828   & 0.87 & \textbf{{0.89}}  & \underline{{0.871}}& \color{gray}{0.802} &   \color{gray}{0.858} \\
Consolidation       & 0.703 & 0.722   & \underline{{0.80}} & 0.79  & \textbf{{0.810}}& \color{gray}{0.774} &   \color{gray}{0.803} \\
Edema               & 0.805 & 0.835   & \underline{{0.88}} & \textbf{{0.91}} & 0.862& \color{gray}{0.813} &  \color{gray}{0.857} \\
Emphysema           & 0.833 & 0.856   & \underline{{0.91}} & \textbf{{0.93}} & 0.896& \color{gray}{0.757} &  \color{gray}{0.879} \\
Fibrosis            & 0.786 & \underline{{0.803}}   & 0.78 & 0.80 & \textbf{{0.849}} & \color{gray}{0.793} &  \color{gray}{0.836} \\
Pleural\_Thickening & 0.684 & 0.749   & 0.79 & \underline{{0.80}} & \textbf{{0.829}} & \color{gray}{0.779} &   \color{gray}{0.805} \\
Hernia              & 0.872 & 0.894   & 0.70 & \underline{{0.92}} & \textbf{{0.927}} & \color{gray}{0.823} &  \color{gray}{0.832} \\
\hline
Mean                & 0.745 & 0.789   & 0.81 & \underline{{0.83}} & \textbf{{0.855}} & \color{gray}{0.798} & \color{gray}{0.833} \\
\hline
\end{tabular}
{\caption{\small \textbf{AUC-ROC score of different disease identification methods.} Our model outperforms state-of-the-art methods, including Wang~\etal.~\cite{wang2017chestx}, CheXNet~\cite{rajpurkar2017chexnet}, Li~\etal.~\cite{li2018thoracic}, and CIA-Net~\cite{liu2019align}. The ablation results (gray) of variants of our model without region verification (RV) and distance learning (DL) are better than vanilla DenseNet-121~\cite{rajpurkar2017chexnet} without these modules. Note that $^{\diamond}$ denotes using additional bounding box supervision and $^{\ast}$ denotes based on our implementation; results for other models are from the papers. Bold and underline indicate the best and second best results, respectively.}
\label{tab:result_cls}}
\vspace{-2mm}
\end{table}

Our end-to-end framework can be applied to various other backbones with modest hyper-parameter tuning such as changing the input image resolution and threshold of CAM. We tested the performance of the vanilla ResNet-50~\cite{he2016deep} network, which achieved 0.770 mean AUC-ROC. When ResNet-50 was used as the backbone in our proposed model, it achieved 0.822 mean AUC-ROC, demonstrating the applicability and effectiveness of distance learning and region verification modules when applied to another popular backbone network.

\paragraph{Ablation studies.}
We show ablation studies in \autoref{tab:result_cls}. We consider the individual contributions of region verification and of distance learning to our model's performance on the right side of \autoref{tab:result_cls} (in gray). Note that use of both modules improves on the performance of one alone, indicating that they model complementary information.
We can further compare the performance of our base model with only distance learning (labeled ``w/o RV'') or only region verification (labeled ``w/o DL'') to the column headed \cite{rajpurkar2017chexnet}, which we re-implemented and which serves as a base model to which the region verification and distance learning modules are added. We can note that each module alone
improves performance over the baseline.

\paragraph{Learning with hard examples.}
In our model in \autoref{tab:result_cls}, the hard positive examples included partial label matches (Section~\ref{sec:triplet}). When positive/negative examples are randomly selected without considering hardness, performance decreases from 0.855 to 0.840, verifying the utility of hard examples. Including partial matches vs.\ identical matches only performed similarly (0.855 vs.\ 0.853), indicating positive examples need not include partial matches.

\subsection{Disease localization}
\paragraph{Main results.}
We compare the localization performance of our model to the three baseline models used in \autoref{tab:loc} that provided quantitative localization results. For the MIL-based models \cite{li2018thoracic,liu2019align}, we compare when the models are trained without any test data; that is, when no disease bounding boxes are used (in cross-validation). The performance of our model was similar to that of the best performing model, Li \etal~\cite{li2018thoracic}, except when T~(IoU) was 0.1, where it was second best. 

\begin{table}[]
\centering
\begin{tabular}{c|l|cccccccc|c}
T (IoU) & Method & Ate. & Car. & Eff. & Inf. & Mas. & Nod. & Pn1 & Pn2 & Mean \\
\shline
\multirow{4}{*}{0.1} & \hspace{.5em}\cite{wang2017chestx} & 0.69 & 0.94 & 0.66 & 0.71 & 0.40 & 0.14 & 0.63 & 0.38 & 0.57 \\
                     & \hspace{.5em}\cite{li2018thoracic}$^{\diamond}$ & \textbf{0.71} & 0.98 &  \textbf{0.87} &  \textbf{0.92} &  \textbf{0.71} &  \textbf{0.40} & 0.60 &  \textbf{0.63} &  \textbf{0.73} \\
                     & \hspace{.5em}\cite{liu2019align}   & 0.39 & 0.90 & 0.65 & 0.85 & 0.69 & 0.38 & 0.30 & 0.39 & 0.60 \\
                     & \hspace{.5em}Ours                &  0.59&	 \textbf{0.99}&	0.85&	0.76&	0.61&	0.23&	 \textbf{0.68}&	0.49&	0.65  \\
                     \hline
\multirow{4}{*}{0.3} & \hspace{.5em}\cite{wang2017chestx} & 0.24 & 0.46 & 0.30 & 0.28 & 0.15 & 0.04 & 0.17 & 0.13 & 0.22 \\
                     & \hspace{.5em}\cite{li2018thoracic}$^{\diamond}$ & 0.36 & 0.94 & \textbf{0.56} & 0.66 & 0.45 & \textbf{0.17} & 0.39 & \textbf{0.44} & \textbf{0.50} \\
                     & \hspace{.5em}\cite{liu2019align}   & 0.34 & 0.71 & 0.39 & 0.65 & \textbf{0.48} & 0.09 & 0.16 & 0.20 & 0.38 \\
                     & \hspace{.5em}Ours                  & \textbf{0.51} & \textbf{0.96} & \textbf{0.56} & \textbf{0.67} & 0.45 & 0.16 & \textbf{0.43} & 0.21 & \textbf{0.50} \\
                     \hline
\multirow{4}{*}{0.5} & \hspace{.5em}\cite{wang2017chestx} & 0.05 & 0.18 & 0.11 & 0.07 & 0.01 & 0.01 & 0.03 & 0.03 & 0.06 \\
                     & \hspace{.5em}\cite{li2018thoracic}$^{\diamond}$ & 0.14 & 0.84 & \textbf{0.22} & 0.30 & 0.22 & \textbf{0.07} & 0.17 & \textbf{0.19} & \textbf{0.27} \\
                     & \hspace{.5em}\cite{liu2019align}   & 0.19 & 0.53 & 0.19 & \textbf{0.47} & \textbf{0.33} & 0.03 & 0.08 & 0.11 & 0.24 \\
                     & \hspace{.5em}Ours                  & \textbf{0.20} & \textbf{0.92} & 0.19 & 0.39 & 0.20 & 0.06 & \textbf{0.18} & 0.04 & \textbf{0.27} \\
                     \hline
\multirow{4}{*}{0.7} & \hspace{.5em}\cite{wang2017chestx} & 0.01 & 0.03 & 0.02 & 0.00 & 0.00 & 0.00 & 0.01 & 0.02 & 0.01 \\
                     & \hspace{.5em}\cite{li2018thoracic}$^{\diamond}$ & 0.04 & 0.52 & 0.07 & 0.09 & 0.11 & \textbf{0.01} & \textbf{0.05} & 0.05 & 0.12 \\
                     & \hspace{.5em}\cite{liu2019align}   & \textbf{0.08} & 0.30 & \textbf{0.09} & \textbf{0.25} & \textbf{0.19} & 0.01 & 0.04 & \textbf{0.07} & \textbf{0.13} \\
                     & \hspace{.5em}Ours                  & 0.04 & \textbf{0.72} & 0.03 & 0.15 & 0.02 & 0.00 & 0.03 & 0.01 & \textbf{0.13} \\
                     \hline
\end{tabular}
\caption{\small \textbf{Accuracy (in \%) of different disease localization methods under various T (IoU).} Using the CAM~\cite{zhou2016learning} weakly-supervised localization method, our model is on par or even surpasses SOTA methods, including Wang~\etal.~\cite{wang2017chestx}, Li~\etal.~\cite{li2018thoracic}, and CIA-Net~\cite{liu2019align}. $^{\diamond}$ denotes using additional bounding box supervision. Pn1: Pneumonia. Pn2: Pneumothorax.}
\label{tab:loc}
\end{table}

\paragraph{Ablation studies.}
We analyze the effect of different modules for disease localization. 
In~\autoref{tab:ablation_loc}, we show ablation results of full models without distance learning and region verification, respectively. Note that each of the modules does contribute to the overall performance of the model, and that overall, region verification contributed more strongly. This stronger contribution may be expected since region verification tends to insure that the identified region contains the features of the targeted disease.

\begin{figure*}[t]
    \centerline{\includegraphics[width=1\linewidth]{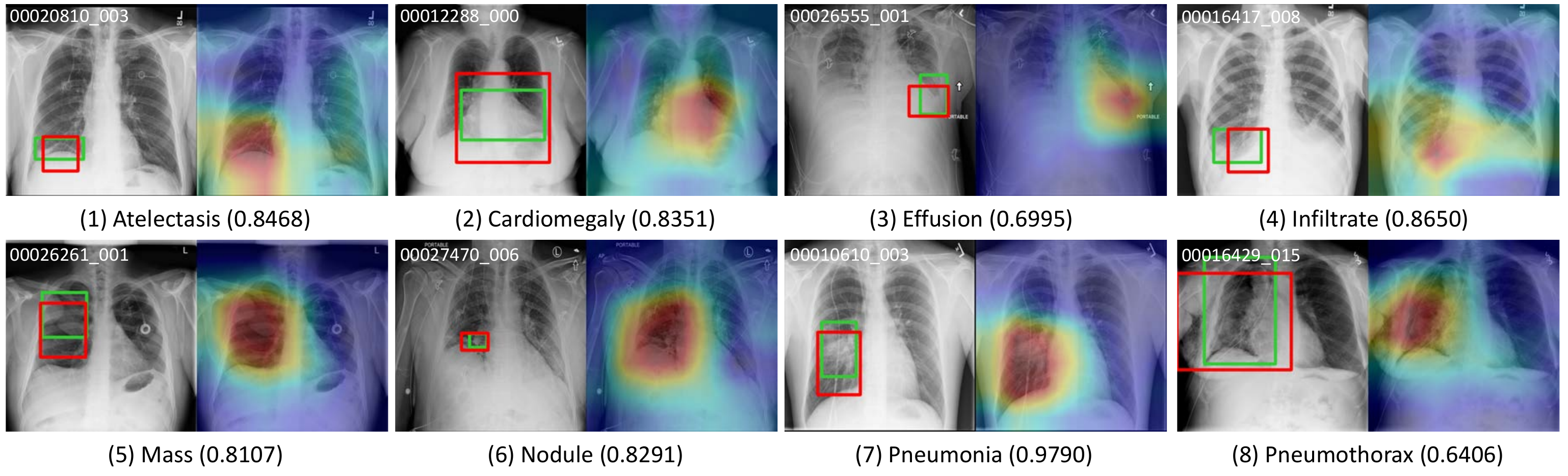}}
    \caption{\small \textbf{Qualitative results of 8 disease types.} For each sample, we show the original image, CAM-based~\cite{zhou2016learning} heat map, localization result as well as confidential score. Green and red boxes indicate ground truth and predicted result, respectively. Image IDs are in the top-left corner.}
    \label{fig:qualitative}
  \vspace*{-2mm}
\end{figure*}

\begin{table}[]
\centering
\begin{tabular}{c|c|c|c|c}
Model                   & IoU (0.1) & IoU (0.3) & IoU (0.5) & IoU (0.7) \\
\shline
Full model              &   0.65        & 0.50      & 0.27      & 0.13      \\
w/o DL   &    0.62       & 0.49      & 0.27      & 0.06      \\
w/o RV &     0.60      & 0.48      & 0.24      & 0.10     \\
\hline
\end{tabular}
\caption{\small \textbf{Ablation studies} on distance learning (DL) and region verification (RV) modules for disease localization. We show overall accuracy with different thresholds.}
\label{tab:ablation_loc}
\vspace{-2mm}
\end{table}

\paragraph{Qualitative results.}
We show qualitative results in~\autoref{fig:qualitative} for one example of each of the 8 diseases with location labels. We observe for these images that the probability of the disease, shown next to the disease name, indicates that the disease is present and that  the high activations in the heatmap are aligned with the ground truth disease locations.

\section{Discussion}
\label{sec:disc}
In this paper we proposed the use of distance learning and region verification for disease identification and localization. 
We used disease-specific region features which provide a variable size context. Our ablation studies indicate that this strongly improves disease identification and localization performance.
This strong contribution may be expected since region verification tends to insure that the identified region contains the features of the targeted disease. We also investigated the use of distance learning with hard example mining for a multi-label task. We observed that use of triplets containing hard examples improves performance, and that the positive examples can be selected from either identical  disease label matches only or also include partial label matches.
Our experiments showed that each method individually leads to performance improvement, and together offer state-of-the-art performance for disease identification and competitive performance for localization.

\paragraph{Acknowledgements} This work was primarily done when Cheng Zhang was a research intern at FX Palo Alto Laboratory (FXPAL). He would like to thank colleagues from FXPAL for the collaboration, advice and for providing an open and inspiring research environment.

{\small
\bibliography{main.bib}
}
\end{document}